\documentclass[conference]{IEEEtran}
\IEEEoverridecommandlockouts
\usepackage{cite}
\usepackage{amsmath,amssymb,amsfonts}
\usepackage{algorithmic}
\usepackage{graphicx}
\usepackage{textcomp}
\usepackage{censor}
\usepackage{xcolor}
\def\BibTeX{{\rm B\kern-.05em{\sc i\kern-.025em b}\kern-.08em
    T\kern-.1667em\lower.7ex\hbox{E}\kern-.125emX}}

\usepackage{xurl} 
\usepackage[colorlinks,allcolors=blue]{hyperref} 

\newcommand{\frombody}[3]{\noindent\textcolor{#1}{{$\bf [\![\!\![$}\underline{\scshape{#2}:} \textsl{#3}{$\bf ]\!\!]\!]$}}}

\renewcommand{\frombody}[3]{}

\usepackage{adjustbox}
\usepackage{array}
\usepackage{hyperref}

\begin{document}

\StopCensoring

\title{The Perceived Danger (PD) Scale:\\Development and Validation\\

\thanks{\blackout{This work was supported in part by NRL to GT. The
  views and conclusions contained in this document are those of the
  authors and should not be interpreted as necessarily representing
  the official policies, either expressed or implied, of the US
  Navy.}}
}


\author{\IEEEauthorblockN{Jaclyn Molan}
\IEEEauthorblockA{\textit{Department of Psychology} \\
\textit{George Mason University}\\
Fairfax, Virginia, USA \\
jackiemolan@gmail.com}
\and
\IEEEauthorblockN{Laura Saad}
\IEEEauthorblockA{\textit{NRC Postdoctoral Fellow} \\
\textit{Naval Research Laboratory}\\
Washington, DC, USA \\
laura.s.saad.ctr@us.navy.mil}
\and
\IEEEauthorblockN{Eileen Roesler}
\IEEEauthorblockA{\textit{Department of Pychology} \\
\textit{George Mason University}\\
Fairfax, Virginia, USA\\
eroesle@gmu.edu}
\and[\hfill\mbox{}\par\mbox{}\hfill]
\IEEEauthorblockN{J. Malcolm McCurry}
\IEEEauthorblockA{\textit{Arcfield} \\
Chantilly, Virginia, USA \\
John.Mccurry@arcfield.com}
\and
\IEEEauthorblockN{Nathaniel Gyory}
\IEEEauthorblockA{\textit{Artificial Intelligence Center} \\
\textit{Naval Research Laboratory}\\
Washington, DC, USA \\
nathaniel.p.gyory.civ@us.navy.mil}
\and
\IEEEauthorblockN{J. Gregory Trafton}
\IEEEauthorblockA{\textit{Artificial Intelligence Center} \\
\textit{Naval Research Laboratory}\\
Washington, DC, USA \\
greg.j.trafton.civ@us.navy.mil}
}


\maketitle

\begin{abstract}
There are currently no psychometrically valid tools to measure the perceived danger of robots. To fill this gap, we provided a definition of perceived danger and developed and validated a 12-item bifactor scale through four studies.
An exploratory factor analysis revealed 
four subdimensions of perceived danger: affective states, physical vulnerability, ominousness, and cognitive readiness. 
A confirmatory factor analysis confirmed the bifactor model.
We then compared the perceived danger scale to the Godspeed perceived safety scale and found that the perceived danger scale is a better predictor of empirical data. We also validated the scale in an in-person setting and found that the perceived danger scale is sensitive to robot speed manipulations, consistent with previous empirical findings.
Results across experiments suggest that the perceived danger scale is reliable, valid, and an adequate predictor of both perceived safety and perceived danger in human-robot interaction contexts. 

\end{abstract}


\section{Introduction}
The increasing frequency of interactions between humans and robots necessitates a focus on 
maximizing safety and understanding how perceptions of danger arise.
If people perceive robots to be dangerous, they might think and behave in ways that do not promote effective human-robot interaction (HRI). For example, a home assistance robot that aids in older adult care may be required to perform important tasks that improve the health and well-being of seniors, like helping with meal times by cutting food into manageable portions \cite{smarr2012older}. If the home assistance robot is holding a knife for this purpose and moves too quickly near them, the person may perceive the robot as dangerous \cite{leusmann2023database}. This could cause them to refuse the robot's care or feel heightened levels of stress while in their own home, which could have pronounced adverse effects on both health and the likelihood of wanting to interact with a robot in the future.
The importance of safety (and its counterpoint, danger) is not limited to healthcare domains, so it is generally important to better understand the factors that underlie the perceived danger of robots so that we can improve the safe design of robots in the future.
The present research developed and validated a scale to measure the perceived danger of robots.

While our primary focus is on perceived danger, much of the existing research in HRI has concentrated on perceived safety. Although safety and danger may be distinct constructs, they are likely strongly related. Therefore, we will explore existing literature surrounding both safety and danger. 


Existing HRI scales largely focus on the perceived safety of robots. 
Notable examples include the Godspeed perceived safety (PS) subscale \cite{bartneck2009measurement}; the Robotic Social Attributes Scale (RoSAS; \cite{carpinella2017robotic}), which includes a discomfort subdimension that reflects safety concerns; and the Negative Attitudes toward Robots Scale (NARS; \cite{nomura2006nars}), which includes items concerned with feelings of unease and comfort around robots.
A common thread across these scales is that safety exists as a dimension within a larger construct.
However, perceived safety (or perceived danger) is its own construct and should be measured accordingly.

There are a limited number of previous studies that have aimed to measure perceived danger.
For example, Carr et al. \cite{carr2001exploratory} assessed danger by asking participants how safe, relaxed, vulnerable, threatened, and at risk they felt. Additionally, Young et al. \cite{young1990judgments} found that overall ``unsafeness'' - a composite term for hazardous, risky, dangerous, and hazardous-to-use - and potential severity of injury play the largest role in judgments of whether a person should act cautiously. Although these studies focused on aspects of perceived danger, they did not develop and validate their items to the extent that would constitute a perceived danger scale. In addition, neither of these studies were concerned with perceived danger in the context of HRI. 


It is unclear from the previous literature whether perceived safety and perceived danger are unidimensional or multidimensional constructs. The majority of existing perceived safety and perceived danger measures focus on emotional states, such as how anxious, nervous, and/or worried one feels \cite{sayin2015sound, cao2021development, nomura2006ras, kulic2007pre}. However, other studies have shown that both constructs are comprised of aspects beyond emotional states alone.
For example, several researchers have focused on the potential for physical harm, with items like ``hazardous,'' ``could get hurt easily,'' ``risky,'' and ``fear for health'' \cite{hayes1998measuring, lee2015can, kamide2012new}. Other studies have shown that danger and safety can be partially characterized by how threatening (or safe) a situation or environment is, with items like ``I feel safe when walking alone during the day'' \cite{syropoulos2024safe}. In fact, the Safety Rating Scale \cite{culbertson2001impact} is a comprehensive scale specifically designed to assess location-based aspects.
Finally, some researchers have found that there is an element of cognitive readiness associated with perceived safety and danger. For example, previous studies have found that confidence, feeling like one has control over the potentially dangerous entity, and the frequency at which entities follow or violate safety rules are factors that contribute to perceived safety and danger \cite{syropoulos2024safe, akalin2022you, cao2021development, lee2015can}. Being alert for possible dangers is a standard hallmark of dangerous situations, so it is important to consider cognitive readiness when evaluating perceived danger.
Collectively, these studies suggest that there are numerous additional facets, beyond emotional states, that are likely to be part of perceived danger.


Given the ga in the literature, we aimed to develop a clear, comprehensive definition of perceived danger. A primary consideration was to ensure that danger was not being defined only in terms of safety, as it is common in perceived safety studies that safety is defined in terms of danger. can be found in Bartneck et al.'s \cite{bartneck2009measurement} development of the Godspeed series, where perceived safety is defined as a person's perceived level of danger and comfort during an interaction with a robot. Defining one construct in terms of the other limits the scope of the definition and the scale that follows from it; the Godspeed definition is primarily focused on emotional aspects of perceived safety, which is not conducive to the investigation of the possibility of multidimensionality. Therefore, there is a need for a definition of perceived danger that is broad enough to encompass numerous possible dimensions, including emotional, physical, and cognitive.
Additionally, we aimed to define perceived danger independently of safety because perceived safety is a more passive feeling, as people typically do not consider their feelings of safety until they begin to feel unsafe. Perceived danger, on the other hand, is a more salient and tangible feeling. Therefore, perceived safety and perceived danger are not directly opposite constructs, and lack of one does not necessarily signify an active perception of the other. For these reasons, we sought to develop a definition that focuses on perceived danger as its own construct rather than as it relates to perceived safety.

Keeping the aforementioned considerations in mind, we propose the following definition of perceived danger: \textit{Perceived danger is the anticipation of harmful consequences of the actual or imagined interaction with another entity}. In this report, we are focusing on the perceived danger of robots, though we hope that this scale will be applicable to other entities as well.


To date, there are no psychometrically valid tools in the HRI literature for measuring perceived danger. We aim to fill this gap by developing and validating a perceived danger (PDscale across four experiments. In Experiment 1, we created an initial set of items, conducted exploratory factor analysis (EFA), and used bifactor modeling to show which items were consistent with perceived danger, allowing us to generate a final set of items to measure perceived danger. In Experiment 2, we verified the factor structure and strength of the final set of items using confirmatory factor analysis (CFA). In Experiment 3, we validated the scale using ordinal regression. Experiment 4 was an in-person validation study, where we included a robot speed manipulation to replicate previous empirical findings about variables that affect the perceived danger of robots. The final version of the scale consists of the bolded items in Table~\ref{tab2}. Our aim is to equip the HRI community with a ychometric tool to measure perceived danger, driving safer practices through a deeper understanding of the role of danger in human-robot interactions.

\section{Item Generation}
Item generation took place through an iterative expert process using semantic and pragmatic analyses to create items that were similarly syntactic, yet semantically distinct enough from one another. 
Following a thorough review of the literature, items were first extracted from several relevant papers about perceived safety and perceived danger of robots and other physical entities
\cite{bartneck2009measurement, cao2021development, nomura2006nars, nomura2006ras, carpinella2017robotic, hayes1998measuring}. These initial items were then supplemented with additional items that were generated based on our own definition of perceived danger. To ensure that a wide range of perceived danger was being covered, we proposed the following four a priori categories: affective states, physical vulnerability, perceptions of robot's actions, and cognitive readiness. After undergoing several phases of internal evaluation, we refined the large set of items to 22 items, which can be found in Table~\ref{tab2}.

All studies were approved by the NRL Institutional Review Board, and all participants provided informed consent.

\begin{table*}[tbp]
\caption{Description of Video Stimuli}
\begin{center}
\begin{tabular}{|l|l|l|l|}
\hline
Label & Video Description & Experiment Used & Source \\
\hline
Near Collision & Quadruped robot that rounds a corner and walks toward oncoming human; & 1, 3 & \cite{robots2024collision} \\
 & video ends prior to possible collision & & \\
Arm with Chair & Large robotic arm with chair attached that fli a person uide down & 1, 3 & \cite{robots2024comedian} \\
 & and launches them out of the chair & & \\
Touching Face & Robot with arms that touches a young girl's cheeks when prompted & 1, 3 & \cite{DisneyResearchHub2016handstoface} \\
Grabbing Ball & Robot with arms that tracks the movement of a ball held by a human & 1 & \cite{robots2024grab} \\
 & and grabs the ball once it is in reach & & \\
Hallway Walking & Quadruped robot that rounds a corner and walks down a hallway toward & 2 & \cite{robots2024nosocialnorms} \\
 & oncoming human & & \\
Rotating/Revolving Arm & Large robotic arm with chair attached that spins a person around its base & 2 & \cite{robots2024arm} \\
 & and fli them uide down & & \\
Feeder & Robotic arm holding a fork that picks up a piece of food and brings it & 2 & \cite{robots2024feeder} \\
 & toward a person's mouth & & \\
Handshake & Robot with arms that performs a complex handshake with a person & 2 & \cite{robots2024handshake} \\
\hline
\end{tabular}
\label{tab1}
\end{center}
\end{table*}

\section{Experiment 1}
In order to identify the items that are most closely aligned with perceived danger, we decided to conduct an EFA, a statistical technique used to identify relationshi between variables to reveal the underlying factor structure. 

\subsection{Method}

\subsubsection{Participants}
Prior to any data collection, the target sample size was determined based on previous literature that has suggested that a sample size of at least 300 is needed for EFA \cite{clark1995constructing, guadagnoli1988relation, comrey1988factor}.
A total of 353 participants were recruited through Cloud Research and paid \$1 for their average of four minutes of participation. 13 participants missed the attention check and were 
removed from further analysis, leaving 340 participants. 181 participants were male, 154 were female, two identified as other, and three preferred not to answer. The average age of participants was 41.6 (SD = 12.3) years. 

\subsubsection{Materials}
A total of four videos of robots interacting with humans were collected and pilot-tested prior to this study to ensure a range of perceived danger. Video labels, descriptions, times of use, and citations are provided in Table~\ref{tab1}.


We used 
22  items that were developed in the item generation process (Table~\ref{tab2}). Participants read the prompt ``Answer as if you were the person in the video...'' and rated each item on a 6-point Likert scale ranging from ``not at all'' to ``extremely.'' 

In addition to 
the PD items, we also included an existing scale to investigate the potential relationship between the perceived danger of robots and participants' predisposition toward robots. We used the personal level negative attitude (P-) subscale from the General Attitudes Towards Robots Scale (GAToRS), which is intended to measure anxiety around robots \cite{koverola2022general}. The P- scale 
includes a 7-point Likert response scale ranging from ``strongly disagree'' to ``strongly agree'' and consists of five items (e.g., ``I don't want a robot to touch me'').

\subsubsection{Procedure}
The study was conducted online. Participants first answered a series of demographic questions, read a brief set of instructions for the task, and then answered the GAToRS P- scale. Each participant was randomly assigned to one of the four videos, which they had to watch in full before continuing. At the end of the video, they were taken to a page where they could rewatch the video as desired 
while they 
responded to the  items.
Participants were required to describe the video with at least one sentence and answer all of the  items. After doing so, they had the opportunity to provide experimental feedback. Finally, they were shown a debriefing statement and thanked for their participation.

\subsection{Results}
To determine the number of factors in the  scale, we conducted an EFA. We used a promax rotation based on the assumption of intercorrelation between factors and selected principal axis factoring as the extraction method because it does not require normally distributed data, and our data was non-normal. 
Velicer's MAP test and the comparison data method \cite{ruscio2012determining} both suggested a four-factor solution, so we proceeded. All four factors were interpretable.

The EFA provided sufficient evidence to support the presence of four factors. Results are shown in Table~\ref{tab2}. 
Factor 1, which we call \textit{affective states}, consisted of nine items relating to emotions, including feeling nervous, anxious, and stressed. Factor 2, which we label \textit{physical vulnerability}, was comprised of seven items that were related to the threat of physical harm, including exposure to physical injury and likelihood of the robot to cause pain. Factor 3, which we call \textit{ominousness}, consisted of three items having to do with a sense of threat, such as how intimidating or menacing the robot was. Finally, Factor 4, which we label \textit{cognitive readiness}, was comprised of three items relating to cognitive aspects of danger, such as vigilance and alertness. All factor loadings were $>$ 0.5, and the model accounted for 84\% of the total variance. The scale had very high reliability; $\alpha$ = 0.99.

\begin{table}[tbp]
\caption{Factor Loadings (Experiment 1)}
\begin{center}
\begin{tabular}{|l|c|c|c|c|}
\hline
Item & F1 & F2 & F3 & F4 \\
\hline
\textbf{How nervous would you feel?} & \textbf{1.00} & & & \\
\textbf{How anxious would you feel?} & \textbf{.93} & & & \\
\textbf{How stressed would you feel?} & \textbf{.84} & & & \\
How tense would you be? & .84 & & & \\
How worried would you feel? & .83 & & & \\
How scared would you feel? & .79 & & & \\
How frightened would you feel? & .77 & & & \\
How alarmed would you be? & .52 & & & \\
How concerned would you be? & .50 & & & \\
\textbf{How exposed to physical injury} & & & & \\
\textbf{would you be?} & & \textbf{.92} & & \\
\textbf{How likely was the robot to} & & & & \\
\textbf{cause pain?} & & \textbf{.89} & & \\
\textbf{How severely might you be injured?} & & \textbf{.87} & & \\
How likely was the robot to & & & & \\
cause bodily harm? & & .82 & & \\
How dangerous was the robot? & & .65 & & \\
How vulnerable to harm would & & & & \\
you be? & .30 & .65 & & \\
How hazardous was the robot? & & .62 & & \\
\textbf{How menacing was the robot?} & & & \textbf{.79} & \\
\textbf{How threatening was the robot?} & & & \textbf{.69} & \\
\textbf{How intimidating was the robot?} & & & \textbf{.64} & \\
\textbf{How alert would you be?} & & & & \textbf{.84} \\
\textbf{How vigilant would you be?} & & & & \textbf{.77} \\
\textbf{How cautious would you be?} & & & & \textbf{.56} \\
\hline
Percent variance explained & 34\% & 27\% & 13\% & 10\% \\
\hline
\multicolumn{5}{l}{The prompt ahead of all items was ``Answer as if you were the person} \\
\multicolumn{5}{l}{in the video...'' Factor loadings $<$ 0.3 not reported for clarity.} \\
\end{tabular}
\label{tab2}
\end{center}
\end{table}

The four-factor model was consistent with our expectations of the factor structure. However, all of the factors were highly correlated with one another
(see Table~\ref{tab3}). This led us to consider the possibility of unidimensionality. Interestingly, none of the methods used to determine number of factors suggested a single factor. One approach that can be used when there is evidence for multiple factors (e.g., Velicer) and also evidence for unidimensionality (e.g., very strong correlations between factors) is a bifactor model.  

\begin{table}[tbp]
\caption{Factor Correlation Matrix (Experiment 1)}
\begin{center}
\begin{tabular}{|l|c|c|c|c|}
\hline
 & Factor 1 & Factor 2 & Factor 3 \\
\hline
Factor 2 & .81 & & \\
Factor 3 & .79 & .79 & \\
Factor 4 & .75 & .72 & .59 \\
\hline
\end{tabular}
\label{tab3}
\end{center}
\end{table}

A bifactor model is a multidimensional model in which all items load onto a general factor
and each load onto a given specific factor.
The specific factors are forced to be orthogonal to the general factor and represent residuals relative to the general trait \cite{reise2012rediscovery}. In other words, the specific factors account for unique variance above and beyond the variance accounted for by the general factor \cite{coulacoglou2017psychometrics}. In this case, the general factor is perceived danger and the specific factors are affective states, physical vulnerability, ominousness, and cognitive readiness. The bifactor model is a good fit for our data because it accounts for the components of both unidimensionality and multidimensionality that emerged from the EFA. Bifactor measures of reliability were excellent; $\omega_t$ = 0.99, $\omega_h$ = 0.94. 

We calculated the correlation between the  scale and the GAToRS P- subscale; \textit{r} = 0.38, \textit{p} $<$ 0.001. As expected, this correlation shows that people with a negative predisposition toward robots tend to give higher ratings of perceived danger. The GAToRS P- subscale had high reliability; $\alpha$ = 0.90, $\omega_t$ = 0.91.

\subsection{Discussion}
The results of Experiment 1 showed that the  items load onto four distinct factors (Table~\ref{tab2}), but the factors are highly intercorrelated (Table~\ref{tab3}), suggesting that a bifactor model is most representative of the data. The bifactor model separates the specific factors of affective states, physical vulnerability, ominousness, and cognitive readiness from the general factor of perceived danger.

Evidence suggests that shorter scales are completed more often than longer scales
\cite{hoerger2010participant, liu2018examining}; therefore, our goal was to shorten the scale enough for it to be usable while still keeping adequate coverage of the factors.
We first removed the single item with cross-loadings $\ge$ 0.3.  Additionally, previous researchers have suggested that three is an acceptable  
number of items to retain per factor for maintenance of scale clarity and psychometric soundness \cite{raubenheimer2004item, gogol2014my}. Therefore, we took the three highest loading items from each dimension (bolded in Table~\ref{tab2}) for a total of 12 items to measure perceived danger.

The EFA provided strong evidence that a bifactor model is most representative of the data; however, it is best practice to conduct a CFA on a new independent sample to verify this structure \cite{boateng2018best, brown2015confirmatory}. 




\section{Experiment 2}
CFA is a statistical technique used to confirm whether the relationshi between variables and latent factors fit a pre-determined model of factor structure. The goal of Experiment 2 was to collect data to conduct a CFA in order to ensure that the bifactor model remains the best explanation of the data on the shorter version of the  scale.

\subsection{Method}
\subsubsection{Participants}
Previous literature has suggested a minimum target sample size of 150 participants for conducting a CFA \cite{guadagnoli1988relation}. We recruited 160 participants, one of whom was removed for missing the attention check, leaving 159 total participants for data analysis. All participants were recruited through Cloud Research and paid \$1 for completing the study, which took approximately four minutes on average. 66 participants were male, 92 were female, and one preferred not to answer. The average age of participants was 41.6 (SD = 12.8) years.
For this experiment, as well as all subsequent experiments in this research, participants from previous studies were excluded.

\subsubsection{Materials and Procedure}
Participants observed a new set of four videos (see Table~\ref{tab1}). These videos were 
pilot-tested to ensure a match in the range of perceived danger that was similar to Experiment 1.

A total of 12 items to measure perceived danger were administered to participants (shown in bold in Table~\ref{tab2}). Both the prompt and the Likert scale for rating the items were identical to those used in Experiment 1. The previously used GAToRS P- subscale items \cite{koverola2022general} were also included in this study with the goal of replicating the initial correlational relationship in a new sample. Participants followed the same 
procedure as
Experiment 1.

\subsection{Results}
We conducted CFA on the unidimensional, four-factor multidimensional, and bifactor models to assess the best model to explain the structure of perceived danger and how well our data fit within that structure. To examine how well-fitted each model was, we used the Akaike Information Criterion (AIC). 
Lower AIC scores indicate better model fits, and a difference of two or more denotes statistical significance \cite{wagenmakers2004aic}. 
The AIC of the one-factor model was 5473.3, the AIC of the four-factor model was 5176.0, and the AIC of the bifactor model was 5154.5.
These scores show that the bifactor model fits significantly better than the other models;
therefore, the bifactor model is the best solution for the  scale.


We verified the bifactor model fit using the following CFA maximum likelihood estimation fit indices: comparative fit index (CFI), Tucker-Lewis index (TLI), root mean square error of approximation (RMSEA), and standardized root mean square residual (SRMR). All of these indices showed excellent fit to the data according to well-established cutoffs \cite{hu1999cutoff}; CFI = 0.990, TLI = 0.984, RMSEA = 0.039, SRMR = 0.032.

Reliability of the PD scale was excellent; $\alpha$ = 0.95, $\omega_t$ = 0.97, $\omega_h$ = 0.86. Additionally, we again calculated the correlation between the  scale and the GAToRS P- subscale. Results showed a strong positive correlation, as predicted, indicating that people with a negative predisposition toward robots tend to give higher ratings of perceived danger; \textit{r} = 0.55, \textit{p} $<$ 0.001. The GAToRS P- scale had high reliability; $\alpha$ = 0.91, $\omega_t$ = 0.92.

\subsection{Discussion}
In Experiment 2, we compared three different models (unidimensional, four-factor, and bifactor) using CFA on the reduced set of 12 items to measure perceived danger. 
The CFA results confirmed the results from Experiment 1 that the bifactor model is the best fit for the data.
The general factor can be interpreted as perceived danger and the four specific factors are affective states, physical vulnerability, ominousness, and cognitive readiness. The final version of the  scale contains 12 items and can be found in bold in Table \ref{tab2}.

The bifactor structure has several advantages in interpretation and usage. First, a bifactor model can be used by simply averaging all the items to get a total score. This score can be interpreted as the amount of perceived danger in an environment. Second, the dimensions of the bifactor model can be used or interpreted separately. For example, if a researcher wants to measure only how ominous a robot is, they could use the ominousness subscale.
Our next experiment will attempt to validate the PD scale.

\section{Experiment 3}
The goal of Experiment 3 was to validate the  scale. To do so, we compared it to a related, established scale: the Godspeed PS scale \cite{bartneck2009measurement}. Specifically, we attempted to use each scale as a predictor of how people rank scenarios involving robots displaying different degrees of dangerousness. Our  scale and the Godspeed scale were then compared to determine which fits the data better.


Experiment 3 was divided into two parts.
In Experiment 3a, participants watched a set of videos, rated each one on both the  and Godspeed PS scales, and ranked the robots in terms of perceived danger. 
Since the Godspeed scale is a measure of safety, not danger\footnote{The Godspeed scale is called the PS scale, but because it is a semantic scale with both safety and danger anchors, it could be considered a hybrid scale.  Nonetheless, we take the name of the scale, "Safety," seriously, so both 3a and 3b were important.}, we conducted a second study to evaluate whether  scale could also predict safety ratings.
Therefore, in Experiment 3b, participants ranked the stimuli in terms of perceived safety.

\subsection{Method: Experiment 3a}

\subsubsection{Participants}
We conducted a Monte Carlo power analysis and found that at least 80 participants were necessary to have an 80\% chance of finding a significant effect for each comparison. A total of 114 participants were recruited through Cloud Research, 10 of whom were removed from further analysis for missing the attention check, leaving 104 participants. 48 participants were male and 56 were female. The average age of participants was 41.9 (SD = 13.1) years. They were paid approximately \$2.50, and the experiment took about 10 minutes to complete.

\subsubsection{Materials}
Three previously used videos that captured the widest range of perceived danger were selected for use in this study (see Table~\ref{tab1}).

To assess perceived danger, we used the final version of our  scale (bolded in Table~\ref{tab2}). To assess perceived safety, we used the Godspeed PS scale \cite{bartneck2009measurement}, which 
is prefaced with the prompt ``Please rate your emotional state on these scales:'' and consists of the following three items rated 
on a 5-point semantic differential scale: Anxious - Relaxed, Agitated - Calm, and Quiescent - Surprised.
We again included the GAToRS P- subscale \cite{koverola2022general}. 

\subsubsection{Procedure}
The procedure followed the same sequence of tasks as Experiments 1 and 2, with a few differences. First, each participant watched all three videos that were included in this experiment. Videos were watched one at a time, with the participant answering all of the  and Godspeed PS items about that video before moving on to the next video. Video order was randomized, and the order of  and Godspeed items was counterbalanced across participants. Additionally, after watching and rating all three videos, participants were asked to rank each video from least to most dangerous. They completed this task by dragging a thumbnail of each video to the desired ranking position. Participants were allowed to rewatch the videos while on this page.

\subsection{Results: Experiment 3a}
We first calculated the correlation between the PD scale and the Godspeed PS scale. We expected a high negative correlation because these scales are intended to measure opposing constructs. Results confirmed this prediction; \textit{r} = -0.70, \textit{p} $<$ 0.001. Reliability of the  scale was excellent; $\alpha$ = 0.98, $\omega_t$ = 0.99, $\omega_h$ = 0.93. The Godspeed PS scale had acceptable reliability; $\alpha$ = 0.82, $\omega_t$ = 0.85.

\begin{figure}[tbp]
\centerline{\includegraphics[scale=0.53]{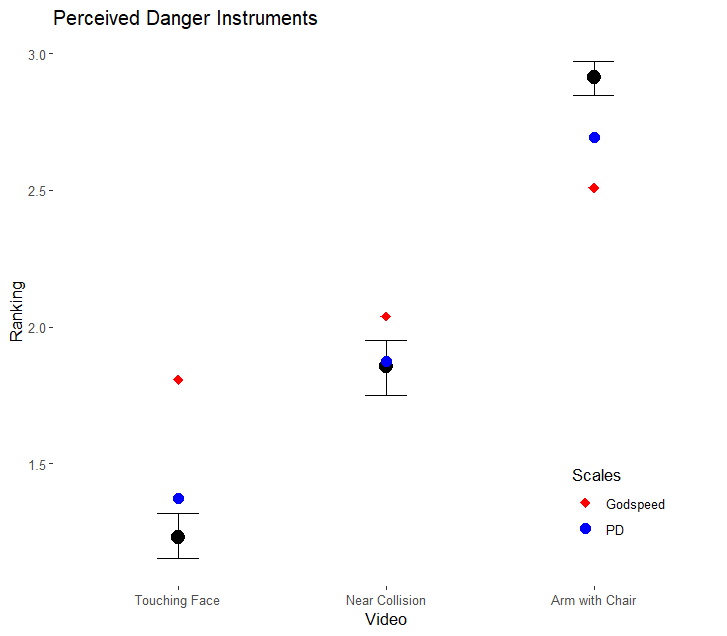}}
\caption{Ordinal regression models for each scale and empirical rankings. Videos were ranked from least (1) to most (3) dangerous. Red diamonds represent the Godspeed PS scale, blue circles represent the  scale, and black circles represent empirical data with a 95\% CI.}
\label{fig1}
\end{figure}

We conducted an ordinal regression for each scale to determine which scale was better at predicting the rank order, from least to most dangerous, of the robots in the videos. Both models were significantly better than chance; \textit{p} $<$ 0.05. The ordinal regression models and the empirical ranking data are shown in Fig.~\ref{fig1}. The AIC value of the PD model was $>$ 2 units less than the AIC of the Godspeed PS model, indicating that the  model was a better predictor of the empirical data \cite{wagenmakers2004aic}; PD AIC = 511.1, Godspeed PS AIC = 607.4. 

We again calculated the correlation between the  scale and the GAToRS P- subscale.
Results showed a moderate correlation, \textit{r} = 0.24, \textit{p} $<$ 0.001, replicating the observed relationship in Experiments 1 and 2. Reliability of the GAToRS P- scale was high; $\alpha$ = 0.87, $\omega_t$ = 0.91.

\subsection{Discussion}
The ordinal regression results support the conclusion that the  scale fits the empirical data significantly better than the Godspeed PS scale.
Although the PD scale was better at predicting the rank ordering of each stimulus in terms of danger, it is possible that the Godspeed scale was not a good predictor of the data because it measures a different construct, safety.
Therefore, 
we decided to run Experiment 3b, where we asked participants to rank each stimulus based on safety. 

\subsection{Method: Experiment 3b}
\subsubsection{Participants}
A total of 127 participants were recruited through Cloud Research and paid \$2.50 for their participation. 3 participants missed the attention check, leaving 124 participants for further data analysis. 50 participants were male, 73 were female, and 1 preferred not to answer. The average age of participants was 43.0 (SD = 13.0) years. The experiment took approximately 10 minutes to complete.

\subsubsection{Materials \& Procedure}
Materials and procedure were the same as those used in Experiment 3a. 
The only difference was that participants were asked to rank the robot in each video from least to most safe.

\subsection{Results: Experiment 3b}
As in Experiment 3a, we calculated the correlation between the PD scale and the Godspeed PS scale and again expected a high negative correlation. Results were aligned with this expectation; \textit{r} = -0.69, \textit{p} $<$ 0.001. Reliability of the  scale was excellent; $\alpha$ = 0.97, $\omega_t$ = 0.98, $\omega_h$ = 0.93. Reliability of the Godspeed PS scale was acceptable; $\alpha$ = 0.71, $\omega_t$ = 0.78.

\begin{figure}[tbp]
\centerline{\includegraphics[scale=0.53]{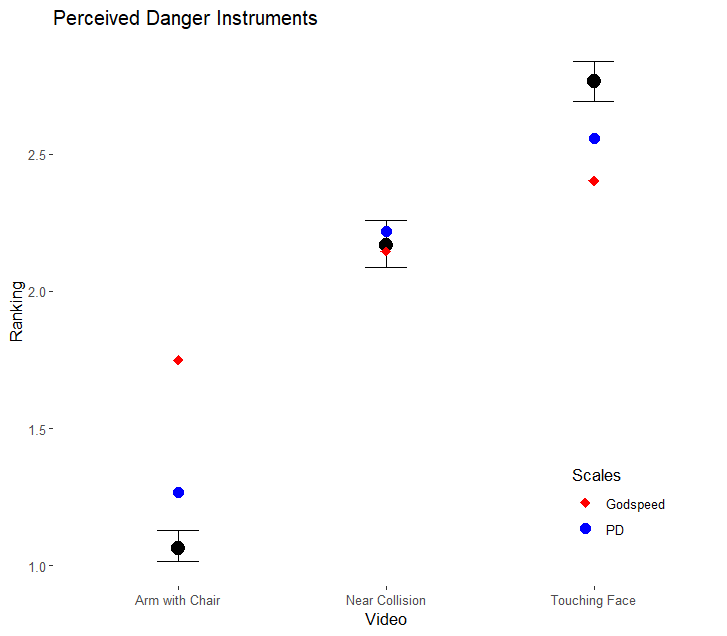}}
\caption{Ordinal regression models for each scale and empirical rankings. Videos were ranked from least (1) to most (3) safe. Red diamonds represent the Godspeed PS scale, blue circles represent the  scale, and black circles represent empirical data with a 95\% CI.}
\label{fig2}
\end{figure}

We conducted an ordinal regression for each scale to determine which scale was better at predicting the rank order, from least to most safe, of the robots in the videos. Both models were significantly better than chance; \textit{p} $<$ 0.05. The ordinal regression models and the empirical ranking data are shown in Fig.~\ref{fig2}. The AIC value of the PD model was $>$ 2 units less than the AIC of the Godspeed PS model, indicating that the PD model was a better predictor of the empirical data \cite{wagenmakers2004aic}; PD AIC = 600.4, Godspeed PS AIC = 732.3. 

We again calculated the correlation between the P scale and the GAToRS P- subscale and found a moderate correlation, \textit{r} = 0.23, \textit{p} $<$ 0.001, replicating the observed relationship in Experiments 1 and 2. 
Reliability of the GAToRS P- scale was high; $\alpha$ = 0.86, $\omega_t$ = 0.91.

\subsection{Discussion}
In Experiment 3, we validated the  scale by comparing it to the Godspeed PS scale using correlation and ordinal regression modeling. Correlation results confirmed the convergent validity of the PD scale, and ordinal regression results showed that the PD scale was better at predicting the rank ordering of the video stimuli in terms of both danger and safety.
Although safety and danger are different constructs, it is apparent that there may be some overlap. This experiment shows that the  scale 
can adequately measure both constructs. Importantly, it performs better than the most commonly employed perceived safety scale in HRI.

\section{Experiment 4}
Experiments 1-3 used video data to create, confirm, and validate the  scale. However, most researchers would want to understand how individuals perceive danger in an embodied context. Previous researchers have shown that the faster a robot moves, the more dangerous it is perceived \cite{kulic2007pre}. We therefore aimed to explore whether our  scale can detect differences between robot speeds. We predicted, consistent with \cite{kulic2007pre}, that faster movements would lead to an increase in the perceived danger of the robot.


\subsection{Method}

\subsubsection{Participants}
21 participants were recruited (17 male, 4 female). The mean age was 36.1 (SD = 11.4) years. The entire experiment took approximately 10 minutes to complete.

\subsubsection{Materials}


We used the quadruped Boston Dynamics Spot robot equipped with the Spot Arm.
It can be difficult to experimentally manipulate perceived danger while keeping participants and observers completely safe. Our primary goal with this experiment was to make sure participants would suffer no harm. Thus, we implemented three different methods to keep participants safe.
First, Spot has built-in obstacle avoidance which causes the robot to move autonomously out of the path of any detected obstacles. 
Second, we used a 
distance verification check; the participant was 4 m away from the starting location, and Spot was not allowed to travel more than 3 m.
In addition, the robot's movement command was attached to a timer. This timer would halt the robot's movement if it continued moving past the expected time of arrival at the participant's observation location. Third, 
an emergency stop button was kept ready during all trials. 

The dependent measures included the final version of the  scale (with modifications to verb tense, see Table~\ref{tab6}), as well as the GAToRS P- subscale \cite{koverola2022general}.

\subsubsection{Procedure}
All participants completed the GAToRS P- scale \cite{koverola2022general} before observing the robot.
At the start of the experiment, participants were told they would be observing a robot and would be asked some questions about their experience. 
Because we needed to keep participants psychologically safe, we told them about the fail-safes described above.
Participants observed the robot grab and remove a screwdriver from a pegboard, turn around, walk toward the participant, and then drop the screwdriver onto the ground approximately 1 m in front of them. Participants remained standing throughout the experiment. 

After the robot dropped the screwdriver, the participant was asked to complete the  scale. The tense of some of the items from the scale was edited so that they were more appropriate for the in-person context. For example, 
``how severely might you be injured?'' was changed to ``how severely might you have been injured?'' The complete list of items used in the in-person experiment can be found in Table~\ref{tab6}. Some items did not need to be customized (bolded items in Table~\ref{tab6}).

\begin{table}[tbp]
\caption{ Items Used in Experiment 4}
\begin{center}
\begin{tabular}{l}
\hline
How nervous did you feel?\\
How anxious did you feel?\\
How stressed did you feel?\\
How exposed to physical injury were you?\\
How severely might you have been injured?\\
\textbf{How likely was the robot to cause pain?}\\
\textbf{How menacing was the robot?}\\
\textbf{How threatening was the robot?}\\
\textbf{How intimidating was the robot?}\\
How alert were you?\\
How vigilant were you?\\
How cautious were you?\\
\hline
\multicolumn{1}{l}{The prompt ahead of all items was} \\
\multicolumn{1}{l}{``During your time with the robot...''} \\
\end{tabular}
\label{tab6}
\end{center}
\end{table}

We varied the walking speed across the two conditions. In the \textit{slow} condition, the robot walked at a speed of 0.5 m/s and dropped the screwdriver at a speed of 0.5 m/s. In the \textit{fast} condition, the robot walked at a speed of 2 m/s and dropped the screwdriver at 2.5 m/s. All other aspects of the trials were the same between conditions. Each participant completed two trials. The order of conditions was counterbalanced between participants.

Note that given our priority on physical and psychological safety of our participants, the instructions and the experiment were extremely safe (i.e., we told participants they would not be in danger and the robot was never on a collision course with the participant). Thus, we expected that the PD scores would differ, but the differences would be small and the scores would be on the lower end of the scale.


\subsection{Results}

A paired-samples t-test revealed an overall difference between the perceived danger in the slow and fast walking conditions; t(20) = -2.50, \textit{p} = 0.02. Specifically, participants perceived the robot to be more dangerous in the \textit{fast} condition (M = 2.37, SD = 1.15) compared to the \textit{slow} condition (M = 2.09, SD = 1.05). As expected, the difference was small, and scores were on the lower end of the scale.


We again calculated the correlation between the PD scale and the GAToRS P- subscale and found a moderate correlation, though the relationship was not significant; \textit{r} = 0.19, $n.s.$
This lack of significance might have been due to the fact that the range of perceived danger was quite low.
Reliability of the GAToRS P- scale was acceptable; $\alpha$ = 0.79, $\omega_t$ = 0.80. The reliability of the  scale was high; $\alpha$ = 0.90, $\omega_t$ = 0.96, $\omega_h$ = 0.67.

\subsection{Discussion}
In Experiment 4, we further validated the  scale by using it in an in-person setting and testing whether it was able to discern differences in robot speed. Based on previous empirical findings \cite{kulic2007pre}, we predicted that a faster-moving robot would be perceived as more dangerous than a slower-moving robot. Results confirmed this prediction, demonstrating that the  scale is suitable for in-person use and can appropriately detect changes in perceived danger.
As predicted, the difference in perceived danger between each condition was small due to our priority of maintaining the physical and psychological safety of participants. It is a strength of the PD scale that it is sensitive enough to capture differences, even in low-danger situations.





\section{General Discussion}
The goal of the present research was to develop and validate a ychometric scale to measure the perceived danger of robots. We first defined our construct and generated an initial set of items based on prior research that reflected our definition. Experiment 1 showed that the best-fitting model for the  scale was a bifactor model with four specific factors: affective states, physical vulnerability, ominousness, and cognitive readiness. We retained three items from each of these factors for a final set of 12 items. In Experiment 2, we confirmed, with strong fit metrics, that the bifactor model was the best fit for the scale. We validated the  scale in Experiment 3 and demonstrated that the  scale fit rank order preferences of safety and danger better than the Godspeed PS scale \cite{bartneck2009measurement}.
Experiment 4 replicated previous research \cite{kulic2007pre} that showed that the faster a robot moved, the more dangerous it appeared. Our  scale showed that people perceived the faster-moving robot to be more dangerous than the slower-moving robot.

\subsection{Theoretical and Practical Implications}
Although perceived danger and perceived safety are certainly related constructs, we believe they are not direct opposites. Safety is a more passive feeling, while danger is a more active, salient feeling. 
Previous research has noted that most definitions of perceived safety have rather defined the \textit{absence} of safety \cite{akalin2022you}, with the most frequently used terms to describe safety being stress, fear, anxiety, and surprise \cite{rubagotti2022perceived, kulic2007pre, cao2021development, hayes1998measuring}.
Danger, as a construct, is more important for practical applications in HRI contexts, but until now, there have been no ychometrically valid tools for measuring the perceived danger of robots. The Godspeed PS scale \cite{bartneck2009measurement} has been the most widely used scale, but we have demonstrated that our  scale is a more effective tool for valid and reliable measurements.

The bifactor model allows for flexible usage of the  scale; the entire scale can be administered to measure a person's overall feeling of danger, or each dimension can be administered as its own subscale to measure more specific aspects of perceived danger. The affective states dimension is concerned with a person's emotions in a situation with a robot. Therefore, it could be useful for populations whose mental health might be affected by their interactions with a robot, such as those receiving healthcare from a robot \cite{riek2017healthcare}. The physical vulnerability dimension focuses on susceptibility to pain and injury, which 
could be applicable to people who work in settings where industrial manufacturing robots are employed. It might be beneficial for a company to understand how the people who work alongside these robots perceive their vulnerability to physical harm in order to develop safety practices that mitigate the potential for harm \cite{zanchettin2015safety}. The ominousness dimension is concerned with the sense of threat that a person feels from a robot. This might be useful for a robot developer who is interested in ensuring that the design of a customer service robot in a hospitality setting does not make people feel threatened or intimidated \cite{lu2019developing}. Finally, the cognitive readiness dimension concentrates on one's feelings of alertness and preparedness to act in a situation with a robot. This subscale might be useful in military contexts, as military robots are often used for potentially hazardous tasks where a certain level of situational awareness by operators and nearby personnel is critical \cite{riley2016situation}. Overall, the bifactor structure is a strength of the  scale, as it will allow researchers to use all or parts of the scale based on their contextual needs.

\subsection{Limitations and Future Directions}
There are some limitations of the present research that are worth noting. First, each experiment used a relatively small selection of videos, so the complete range of perceived danger might not have been captured. The videos also could have included a wider variety of robot morphologies. 

Although the PD scale was designed based on HRI, we believe that it will be applicable to other contexts where perceived danger is important (with modifications to the prompt and/or verb tense as needed). For example, designers of autonomous vehicles would benefit from understanding how people perceive danger while in or when being approached by these vehicles. Future research could focus on whether the  scale can be adequately applied to these types of HRI-adjacent situations.

We should note that perceived (subjective) danger is not the same as actual (objective) danger.
There are many studies that have shown that sometimes safety can increase risky behavior. For example, when new drivers are provided with evasive driving techniques (skid training in particular), crash rates go up, suggesting that they may engage in riskier behavior due to a heightened sense of safety from the training \cite{katila1996conflicting}.
Importantly, actual danger should be considered in relation to our measurement of perceived danger, as the two do not perfectly align.

It is our hope that the PD scale we have developed in this paper will allow other researchers to accurately measure perceived danger, strengthen the community's understanding of danger as its own construct, and inform the implementation of safer practices in HRI.

\section*{Acknowledgment}

\blackout{
The authors thank Beth Philli, Toey Pithayarungsarit, Anthony Harrison, Sangeet Khemlani, William Adams, Ravenna Thielstrom,
Branden Bio, and Ed Lawson for
comments on this research.
}





\newpage
\bibliographystyle{ieeetr}
\bibliography{PD}
\vspace{12pt}

\end{document}